\documentclass[10pt,twocolumn,letterpaper]{article}
\usepackage[accsupp]{axessibility}
\usepackage{cvpr}              

\usepackage{float}
\usepackage{graphicx}
\usepackage{multirow}
\usepackage{wrapfig}
\newcommand{\myparagraph}[1]{
\vspace{0.1cm}\noindent
\textbf{#1.}
}

\usepackage{amsmath}
\usepackage{amsthm}
\newtheorem{property}{\bf Property}[section]
\usepackage{tcolorbox}         
\usepackage{lipsum}            
\newcommand\blfootnote[1]{%
	\begingroup
	\renewcommand\thefootnote{}\footnote{#1}%
	\addtocounter{footnote}{-1}%
	\endgroup
}
\definecolor{cvprblue}{rgb}{0.21,0.49,0.74}
\usepackage[pagebackref,breaklinks,colorlinks,allcolors=cvprblue]{hyperref}

\def\paperID{7563} 
\def\confName{CVPR}
\def\confYear{2025}

\title{Image is All You Need to Empower Large-scale Diffusion \\ Models for In-Domain Generation}
\author{
Pu Cao$^{\dagger}$\hspace{0.65cm} Feng Zhou$^{\dagger}$\hspace{0.65cm}Lu Yang$^*$\hspace{0.65cm}Tianrui Huang\hspace{0.65cm}Qing Song\\
Beijing University of Posts and Telecommunications\\
{\tt\small \{caopu, zhoufeng, soeaver, huangtianrui, priv\}@bupt.edu.cn}} 

\begin{document}

\twocolumn[{
\renewcommand\twocolumn[1][]{#1}
\maketitle
\vspace{-5 mm}
\setlength{\fboxrule}{0.5mm}
\begin{center}
    \centering
    \includegraphics[width=1.0\textwidth]{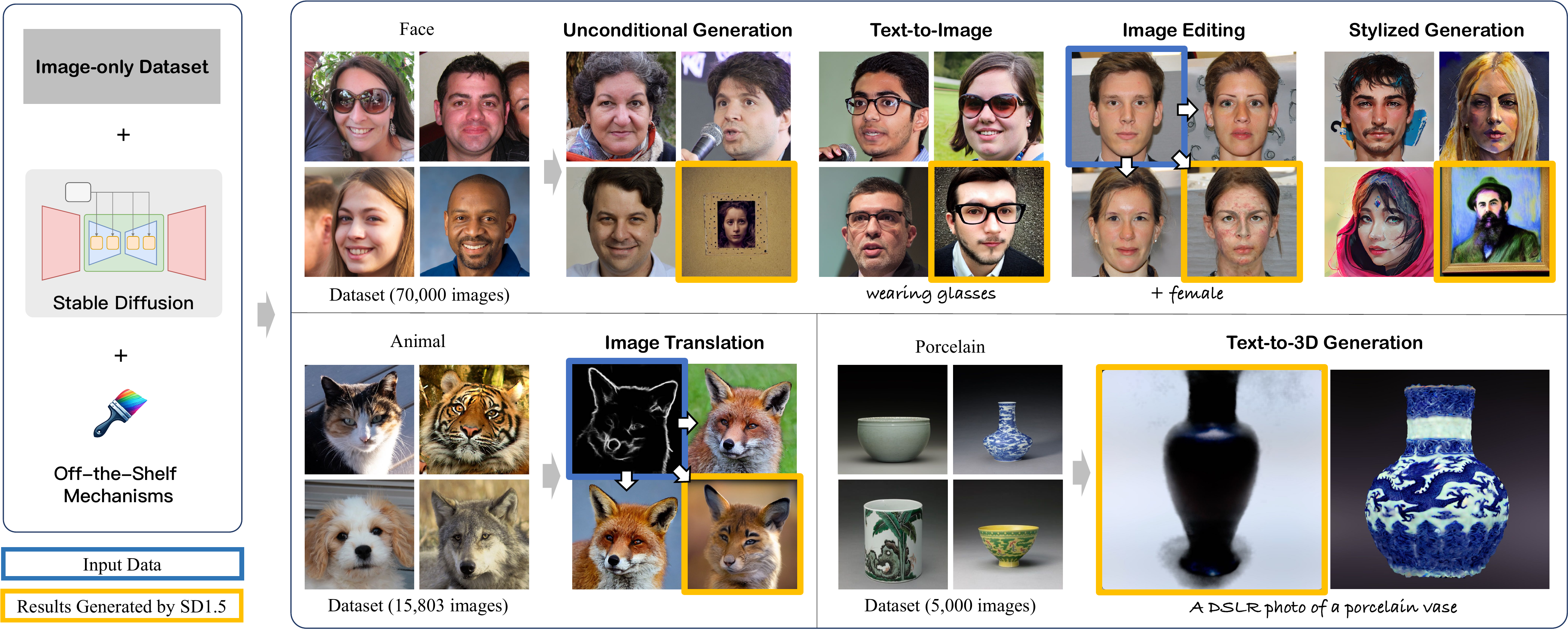}
\vspace{-4mm}
    \captionof{figure}{\textbf{Illutration of In-domain Generation.} In this work, we empower large-scale pre-trained diffusion models using only image data to perform varied generation tasks within each domain with high fidelity and controllability. We mark the input data in blue and the results generated with the original Stable Diffusion v1.5 model in orange.}
    \label{fig:introduce}
\end{center}
}]

\maketitle

\begin{abstract}
\blfootnote{$^\dagger$ The first two authors contributed to this paper equally.}
\blfootnote{$^*$ Corresponding author: Lu Yang.}
\blfootnote{Code is available at \url{https://github.com/PRIV-Creation/In-domain-Generation-Diffusion}.}
In-domain generation aims to perform a variety of tasks within a specific domain, such as unconditional generation, text-to-image, image editing, 3D generation, and more. 
Early research typically required training specialized generators for each unique task and domain, often relying on fully-labeled data. 
Motivated by the powerful generative capabilities and broad applications of diffusion models, we are driven to explore leveraging label-free data to empower these models for in-domain generation.
Fine-tuning a pre-trained generative model on domain data is an intuitive but challenging way and often requires complex manual hyper-parameter adjustments since the limited diversity of the training data can easily disrupt the model's original generative capabilities.
To address this challenge, we propose a guidance-decoupled prior preservation mechanism to achieve high generative quality and controllability by image-only data, inspired by preserving the pre-trained model from a denoising guidance perspective.
We decouple domain-related guidance from the conditional guidance used in classifier-free guidance mechanisms to preserve open-world control guidance and unconditional guidance from the pre-trained model. 
We further propose an efficient domain knowledge learning technique to train an additional text-free UNet copy to predict domain guidance.
Besides, we theoretically illustrate a multi-guidance in-domain generation pipeline for a variety of generative tasks, leveraging multiple guidances from distinct diffusion models and conditions. 
Extensive experiments demonstrate the superiority of our method in domain-specific synthesis and its compatibility with various diffusion-based control methods and applications. 
\end{abstract}

\section{Introduction}
\label{sec:intro}
\begin{figure}
    \centering
	\includegraphics[width=0.5\textwidth]{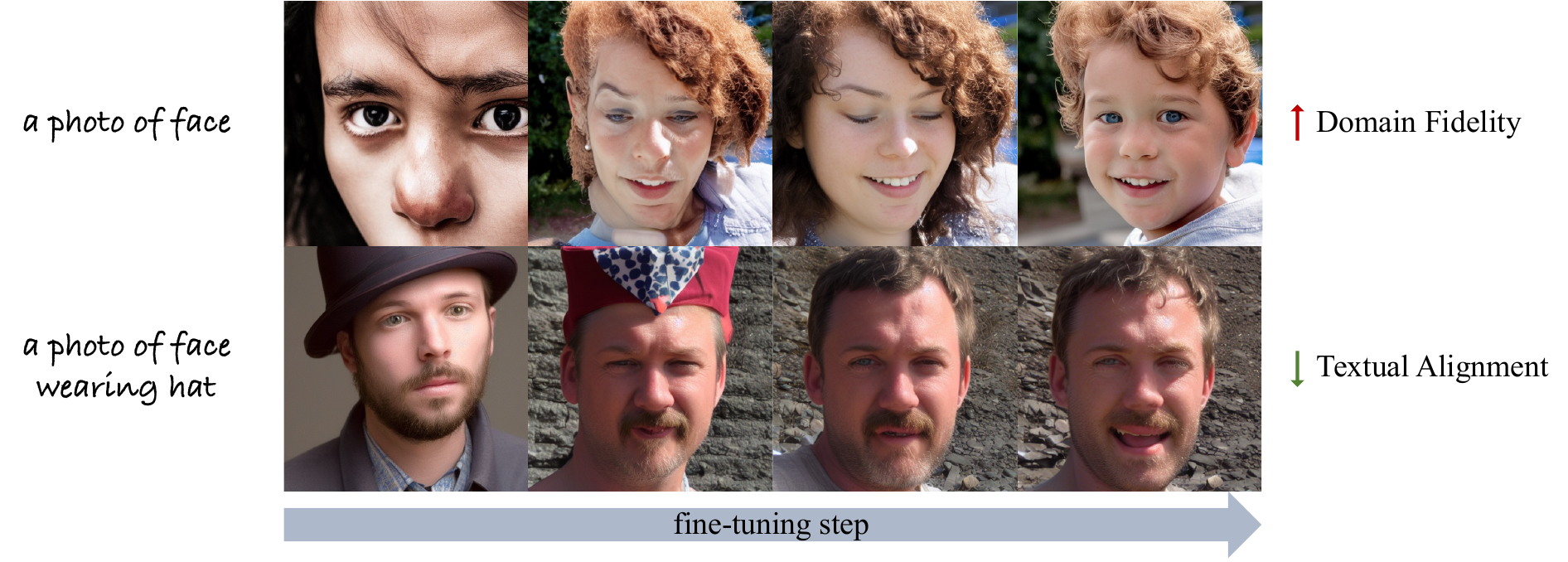}
    \caption{\textbf{Challenge of fine-tuning diffusion models on domain data.} We show the fine-tuning process of Stable Diffusion v1.5 with a facial image dataset (\ie, FFHQ\cite{karras2019style}). The open-world controllability of pre-trained diffusion models is gradually decreased during the fine-tuning process, with domain fidelity improved. 
}
	\label{fig:finetune}
\vspace{-4mm}
\end{figure}

In-domain generation aims to accomplish various tasks within a specific domain, often requiring tailored approaches and labeled data. For instance, pix2pix \cite{isola2017image} was developed for image-to-image translation (\eg, converting segmentation or edge images to realistic images), relying on paired images for training. Similarly, InterfaceGAN\cite{shen2020interfacegan} enables attribute-based image editing but requires attribute-labeled images to train editing vectors. Such traditional methods\cite{isola2017image,chen2019text2shape,fu2022shapecrafter,shen2020interfacegan} depend heavily on specific generative tasks and data domains, which limits their generalization capability.

Thanks to billion-scale image-text datasets~\cite{caesar2018coco,schuhmann2021laion,ramesh2022hierarchical,schuhmann2022laion} and advances in large-scale natural language models~\cite{devlin2019bert,raffel2020exploring,radford2021learning}, recent research on diffusion models has pushed the boundaries of AI-generated content, greatly enhancing both the diversity and controllability of generated outputs \cite{rombach2022high}. With various control mechanisms and approaches (\eg, ControlNet\cite{zhang2023adding}, SDEdit\cite{meng2021sdedit}), diffusion models can also perform diverse generative tasks beyond text-to-image synthesis, alleviating the data demands of traditional generative methods. However, the generated results often fail to fully align with specific data domains, as shown in orange in \cref{fig:introduce}.

While fine-tuning a pre-trained model on domain-specific data is an intuitive way to inject domain knowledge, it is challenging to balance fidelity with controllability, as illustrated in \cref{fig:finetune}. For example, fine-tuning can improve the quality of face generation but may gradually reduce the effect of control prompts like \texttt{wearing hat}.
In this work, we first identify this phenomenon caused by conditional guidance and unconditional guidance catastrophic forgetting during fine-tuning. These two guidances are integrated by classifier-free guidance for improved generative capability, which is employed in most text-to-image diffusion models.
Then, we propose a guidance-decoupled prior preservation method to address these challenges.
We decouple the conditional guidance in CFG into two parts: domain guidance, which is optimized in fine-tuning to enhance fidelity, and control guidance, which is preserved for controllability. 
With introducing multiple guidances in the synthesis process, the control guidance and unconditional guidance, which we want to preserve during personalization, are directly predicted by diffusion priors, while the domain guidance is predicted by an independent diffusion model.
Additionally, we introduce an efficient domain knowledge learning mechanism, designing a null-text Diffusion Model to learn domain knowledge in a concise way.

We conduct extensive experiments to demonstrate the effects of our proposed task and validate the superiority of our approach. 
To summarize, our contributions are as follows:
\begin{itemize}
\item[$\bullet$] We present the task of aligning large-scale diffusion models with specific domains using only image data to perform a variety of generative tasks.
\item[$\bullet$] We propose a guidance-decoupled prior preservation mechanism to address guidance forgetting during fine-tuning, which decouples domain guidance to learn domain knowledge with other guidance preserved. Besides, we propose an efficient learning mechanism to learn domain knowledge.
\item[$\bullet$] We present the pipeline of our proposed method for in-domain generation tasks and conduct experiments and comparisons across multiple domains and tasks to demonstrate the effectiveness and advancement of our approach.
\end{itemize}

\section{Related Works}
\myparagraph{Text-to-image Generation}
As the scale of image-text data grows~\cite{caesar2018coco,schuhmann2021laion,ramesh2022hierarchical,schuhmann2022laion}, large-scale diffusion models~\cite{dhariwal2021diffusion,nichol2021glide,ho2020denoising,rombach2022high} have demonstrated astonishing generative capabilities, able to produce images based on open-world prompts.
Some models, such as Imagen~\cite{saharia2022photorealistic} and Stable Diffusion~\cite{rombach2022high}, have been explored as diffusion priors, applied in downstream tasks like controlled generation~\cite{voynov2023sketch,mou2024t2i,zhang2023adding,li2023gligen} and image editing~\cite{hertz2022prompt,kawar2023imagic,couairon2022diffedit}. For instance, ControlNet~\cite{zhang2023adding} proposes using an additional network to receive more spatial controls to guide generation. Furthermore, diffusion models have demonstrated powerful capabilities in many tasks, such as face parsing \cite{yang2019parsing,yang2020renovating,yang2021quality,yang2022quality,yang2022part,yang2024deep}, style transfer \cite{li2020accelerate,yang2023zero,wang2023stylediffusion}, image manipulation~\cite{dong2023prompt,wu2023latent} and super-resolution~\cite{wu2023hsr,wang2023reconstruct}.

\myparagraph{Fine-tuning Large-scale Diffusion Models}
Some existing methods study learning subjects into large-scale diffusion priors and generate variant images guided by different prompts~\cite{gal2022image,ruiz2023DreamBooth,qiu2023controlling,chen2023disenbooth,liu2023cones,liu2023customizable,arar2023domain,smith2023continual,shi2023instantbooth}. Some methods fine-tune the part of diffusion models with one or a few images of the specific subject~\cite{gal2022image,ruiz2023DreamBooth,qiu2023controlling,chen2023disenbooth}. Textual inversion~\cite{gal2022image} tunes the text embedding of the subject placeholder, while DreamBooth~\cite{ruiz2023DreamBooth} further updates the weight of UNet~\cite{ronneberger2015u}. Some methods further study training-free~\cite{arar2023domain,shi2023instantbooth}, continual customization~\cite{smith2023continual}, or multi-subjects personalization~\cite{liu2023cones,liu2023customizable} and layout-guided generation~\cite{gu2023mix}. However, these methods concentrate on the consistency of given subjects and training efficiency, tailored for the situation with small-scale image datasets and limited training process. 

\myparagraph{Fidelity and Controllability Tradeoff}
In the field of generation, the tradeoff between fidelity and controllability has received widespread attention. In GANs, many inversion methods address this problem through regularization~\cite{roich2022pivotal,cao2022lsap}, latent code discriminator~\cite{tov2021designing}, and invertibility prediction~\cite{parmar2022sam,cao2024decreases}. In the personalization task, thanks to the simple customization process and excellent generalization of diffusion priors, this problem is easily solved by regularization~\cite{ruiz2023DreamBooth}, parameter efficient tuning~\cite{hu2021lora}, orthogonal fine-tuning~\cite{qiu2023controlling}, and so on.
However, a longer training process is required for domain knowledge learning and the existing techniques struggle to solve this problem, highlighting an urgent need for a new approach.

\section{Method}
\label{sec:method}

\subsection{Guidance Catastrophic Forgetting}
\label{sec:task_analysis}

\begin{figure}[t!]
    \centering
	\includegraphics[width=1\linewidth]{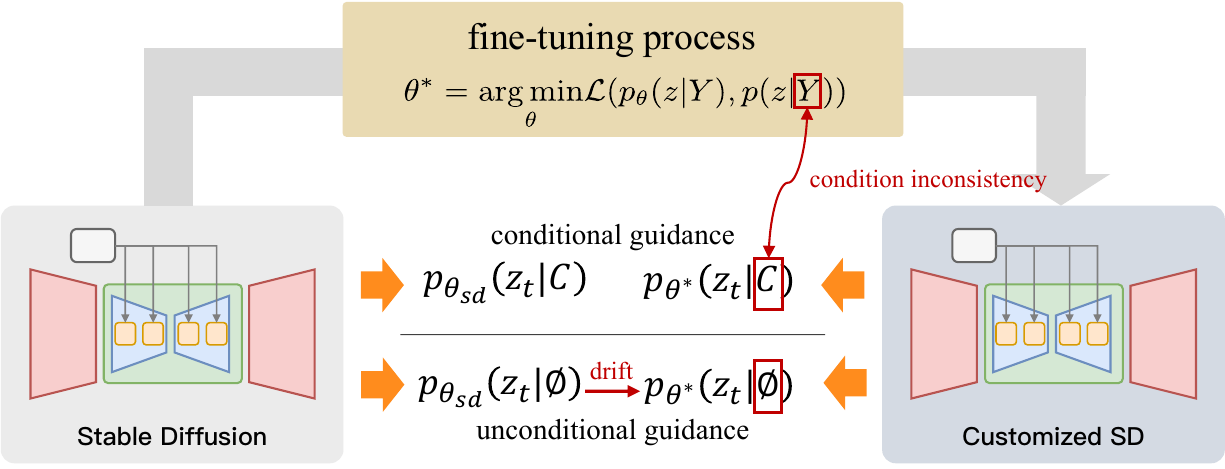}
    \caption{\textbf{Illustration of Fine-tuning Process.} We demonstrate the guidance catastrophic forgetting during fine-tuning process.}
	\label{fig:motivation1}
\end{figure}

\begin{figure}[t!]
    \centering
	\includegraphics[width=0.8\linewidth]{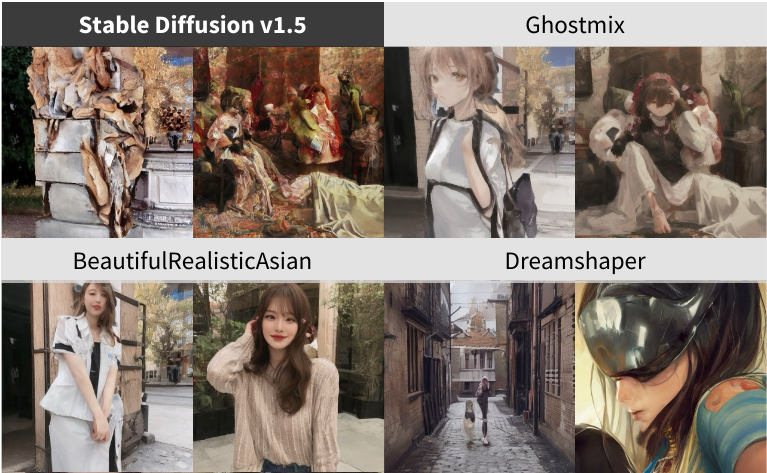}
    \caption{\textbf{Unconditional Guidance Drift.} The unconditional generation results of fine-tuned diffusion models reflect the visual pattern of training datasets, which would cause inaccurate noise estimation.}
	\label{fig:challenge2}
\vspace{-4mm}
\end{figure}

In this section, we delve into the phenomenon of catastrophic forgetting in the training process of diffusion models from the perspective of guidance, allowing us to address this issue at its root.


\myparagraph{Classifier-free Guidance (CFG)} To achieve a better tradeoff between quality and diversity, CFG~\cite{ho2022classifier} is utilized to combine the conditional and unconditional score estimates with a hyperparameter $w$:
\begin{align}
\label{eq:cfg}
\widetilde\epsilon_\theta(z_t,c)=(1+w)\epsilon_\theta(z_t,c)-w\epsilon_\theta(z_t)
\end{align}
In large-scale diffusion model training, suppose that an open-world training dataset denotes as $\{z_i, c_i\}$, where $z_i$ is the training image and $c_i$ is the correspondent text, the training objective is to minimize:
\begin{align}
\label{eq:prior}
{\mathbb{E}}\bigg( & \underbrace{\mathcal{L}(p_\theta(z|c), p(z|c))}_{\text{conditional guidance}} 
+ \underbrace{\mathcal{L}(p_\theta(z), p(z))}_{\text{unconditional guidance}} \bigg)
\end{align}

\myparagraph{Guidance Catastrophic Forgetting} 
During domain data fine-tuning with data $\{z_i, y_i\}$, these two guidances inevitably deviate, which we illustrated in \cref{fig:motivation1}. Notably, $p(z|y)$ represents the probability conditioned on a given domain, which does not imply that text labels of domain data are required. 
For conditional guidance, the condition $y$ in fine-tuning phase is too much simpler than the condition in the pre-trained phase $c$, where the latter is expected to be control generation. For example, although we leverage only facial images in training, we also desire to generate images controlled by text prompts. 
For unconditional guidance, we suppose that the pre-trained model could estimate the accurate $p_{\theta_sd}(z_t|\emptyset)$ since it is trained on open-world data. Hence, fine-tuning on domain data would make $p_{\theta^*}(z_t|\emptyset)$ inaccuracy, illustrated in \cref{fig:challenge2}. The generative results from fine-tuned models reflect the characteristics of the training datasets.



\subsection{Guidance-Decoupled Prior Preservation}
\label{sec:gcfg}
\begin{figure}
    \centering
	\includegraphics[width=0.5\textwidth]{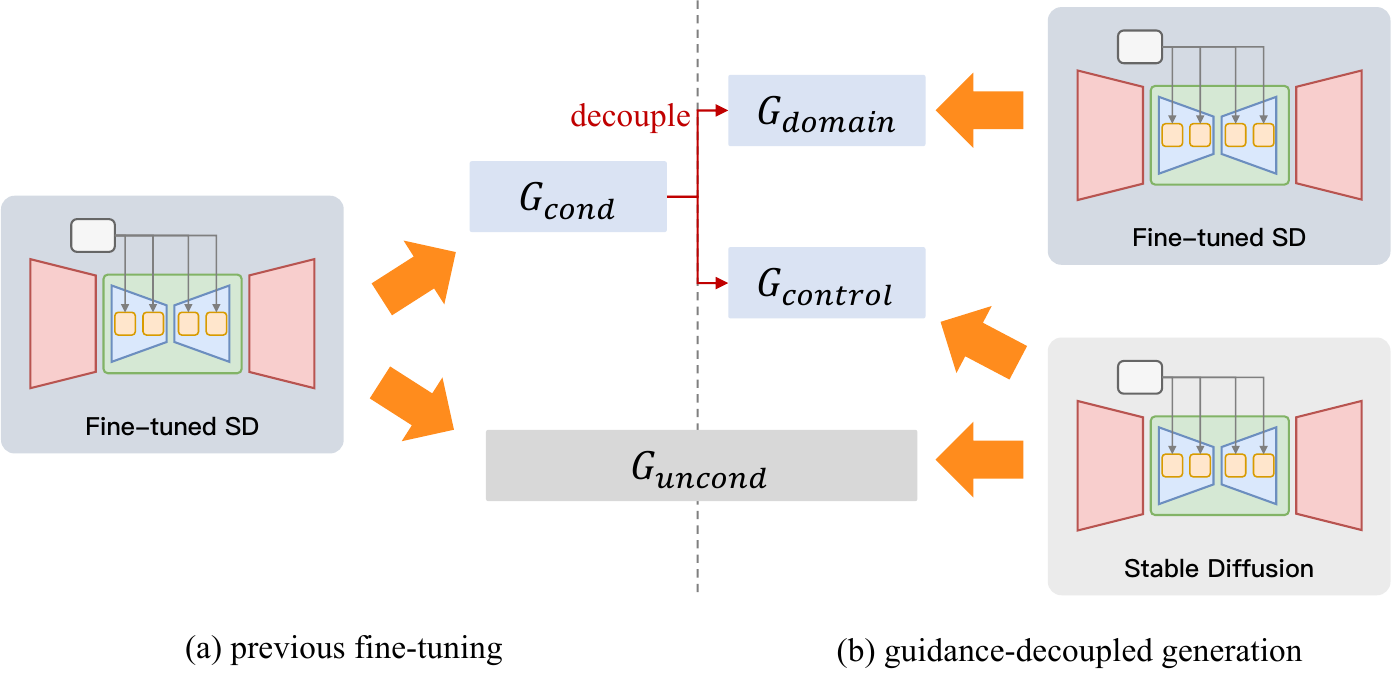}
    \caption{\textbf{Conditional Guidance Decoupling.} 
We compare the guidance estimation between previous customization methods and ours. We decouple the conditional guidance into domain guidance and control guidance while predicting the control guidance and unconditional guidance using the original diffusion model to keep them unchanged.}
	\label{fig:motivation}
\end{figure}


To address the forgetting issue, we propose a guidance-decoupled prior preservation mechanism. We decouple conditional guidance into domain guidance and control guidance, as illustrated in Fig~\ref{fig:motivation}(b).
The domain guidance steers the denoising process to generate domain-aligned images and the control guidance takes charge of handling open-world controls in generation. For example, to generate `a man wearing glasses' within the face domain, domain guidance tells how the human face looks, and control guidance ensures the synthetic faces follow `man' and `wearing glasses'. After such decoupling, control guidance and unconditional guidance can be predicted by diffusion priors and domain guidance can be learned by an additional model.

\myparagraph{Conditional Guidance Decoupling} 
Denoting a sets of conditions as $C=\{c_1,\cdots,c_k\}$ with manually defined intensities $w=\{w_1,\cdots,w_k \}$, we can use multiple UNets $\epsilon_\theta$ to predict conditional and unconditional: $\epsilon_{\theta_0}(z_t)=\nabla_z\log p(z_{t})$ and $\epsilon_{\theta_i}(z_t,c_i)=\nabla_z\log p(z_{t}|c_i)$. The reverse process in each timestamp is as follows:
\begin{align}
\label{eq:gcfg}
\hat\epsilon(z_t,C)=\epsilon_{\theta_0}(z_t)+\sum_{i=1}^K w_i(\epsilon_{\theta_i}(z_t,c_i)-\epsilon_{\theta_0}(z_t))
\end{align}
Hence, we can decouple the original conditional guidance into two components to improve domain alignment and open-world control, respectively. As shown in \cref{eq:gcfg}, we can use different diffusion models to predict each guidance, allowing us to fine-tune a UNet copy to learn domain knowledge and utilize the pre-trained model to control the generation.

However, \cref{eq:gcfg} assumes that all conditions are independent. This assumption leads to the independent maximization of the probability of each condition, which can cause confusion between the conditions. This issue is analyzed further in \cref{sec:appendix_independent}.
Our analysis reveals that a simple yet effective approach to mitigate this issue is to use more detailed descriptions (\eg, \texttt{a photo of a face, wearing glasses}) rather than overly simplified conditions (\eg, \texttt{wearing glasses}). This approach helps reduce confusion and better aligns the guidance with the intended domain.

\myparagraph{Unconditional Guidance Rectification}
Through multi-guidance generation, an intuitive solution to address unconditional guidance drift involves rectifying it using prior diffusion models. We can consider the original pre-trained model as an unconditional guidance predictor, effectively mitigating guidance drift after fine-tuning. This rectification operation can also used in other customized diffusion models, including open-sourced models, as shown in Sec~\ref{sec:discussion}.

\subsection{Efficient Domain Knowledge Learning}
\myparagraph{Null-text Diffusion Model}
\label{sec:cdm}
The additional fine-tuned diffusion model aims to learn the domain guidance by a domain image dataset. To make it concise, we construct a null-text diffusion model to learn domain guidance.

In LDMs, cross-attention is used to fuse text feature and image feature with residual connection, which follows:
\begin{align}
\mathcal{F}_{img}'=\underbrace{W(Softmax(\frac{\mathcal{Q}_{img}\mathcal{K}^T_{text}}{\sqrt{d}})\times\mathcal{V}_{text})}_{text\ feature\ injection} + \mathcal{F}_{img}
\end{align}
where $\mathcal{Q}_{img}$ is transformed from input image feature $\mathcal{F}_{img}$, $\mathcal{K}^T_{text},\mathcal{V}_{text}$ is transformed from text feature, $d$ is the num of channels, and $W$ is a linear projection. 
Since textual information is not used in our fine-tuned model, we can convert the former term into a fixed embedding $E$, and we have: 
\begin{align}
\mathcal{F}_{img}'&=E+\mathcal{F}_{img}
\end{align}

\myparagraph{Initializing $E$ via Textual Features Optimization} 
To better inherent generative capability from diffusion priors, accurate textual features represent a good initializing point in training. Therefore, we draw inspiration from Textual Inversion\cite{gal2022image} and optimize the text features $F^T$, the output of text encoder, and use them as the initial values for $E=W_V(F^T)$. Specifically, we first attain $F^T$ by feeding \texttt{a photo of <domain name>} into the text encoder and optimize $F^T$ by denoising loss. This training process converges in just several minutes, providing a good initial value that accelerates the subsequent fine-tuning process.

\subsection{In-domain Generation with Multi-Guidance}
\label{sec:pipeline}

During domain knowledge learning, we fine-tune a domain diffusion model $\epsilon_{\theta_d}$ to provide domain guidance and employ the Stable Diffusion $\epsilon_{sd}$ to provide preserved guidance, as illustrated in Fig~\ref{fig:motivation}. For control guidance, we can also flexibly adapt existing techniques, including Stable Diffusion, ControlNet, or any open-sourced customized models, to achieve a variety of applications. The reverse process suffices:
\begin{align}
\underbrace{(1-w_d-w_c)\epsilon_{sd}(z_t)}_{{\small unconditional\ guidance}} + \underbrace{w_d\epsilon_{\theta_d}(z_t)}_{{\small domain\ guidance}} 
+ \underbrace{w_c\epsilon_{\theta_c}(z_t,c)}_{{\small control\ guidance}}\nonumber
\end{align}
where $c$ is control signal with model $\epsilon_{\theta_c}$. 
For instance, to generate images conditioned by a canny image in the animal domain, we utilize the diffusion model fine-tuned on an animal dataset as $\epsilon_{\theta_d}$ and the original diffusion model equipped with canny-to-image ControlNet model as $\theta_c$. Experimentally, $w_d$ presents stability in each generative task, similar to the guidance scale used in widely-used diffusion models (\eg, using $w_0=7$ for Stable Diffusion $\epsilon_{sd}$). And we determine the initial value of $w_c$ and $w_d$ by $w_d=\frac{w_0}{2} \cdot \frac{\left\| \epsilon_{sd} \right\|}{\left\| \epsilon_{\theta_d} \right\|}$ and $w_c=\frac{w_0}{2} \cdot \frac{\left\| \epsilon_{sd} \right\|}{\left\| \epsilon_{\theta_c} \right\|}$ to ensure $w_0 \cdot \left\| \epsilon_{sd} \right\| = w_d \left\| \epsilon_{\theta_d} \right\| + w_c \left\| \epsilon_{\theta_c} \right\|$.
We can also employ this generation process in other diffusion-based generative tasks, like image editing and 3D generation.


\begin{figure*}[t!]
	\begin{center}
		\includegraphics[width=1\linewidth]{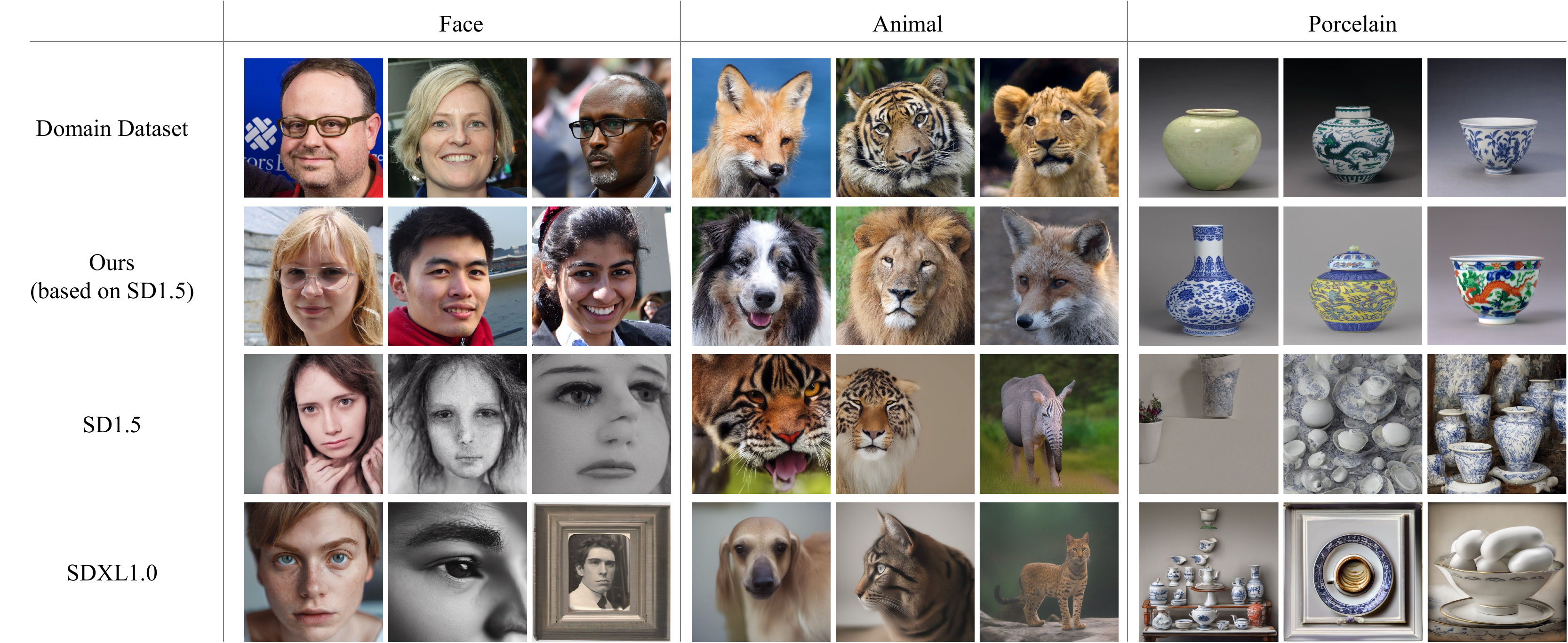}
	\end{center}
	\caption{\textbf{Visualization of Domain Alignment.} Our results are more aligned to given domain dataset. For SD1.5 and SDXL 1.0, we generate images by text \texttt{a photo of <domain name>}.}
	\label{fig:uncond}
\end{figure*}

\begin{figure*}[t!]
	\begin{center}
		\includegraphics[width=1\linewidth]{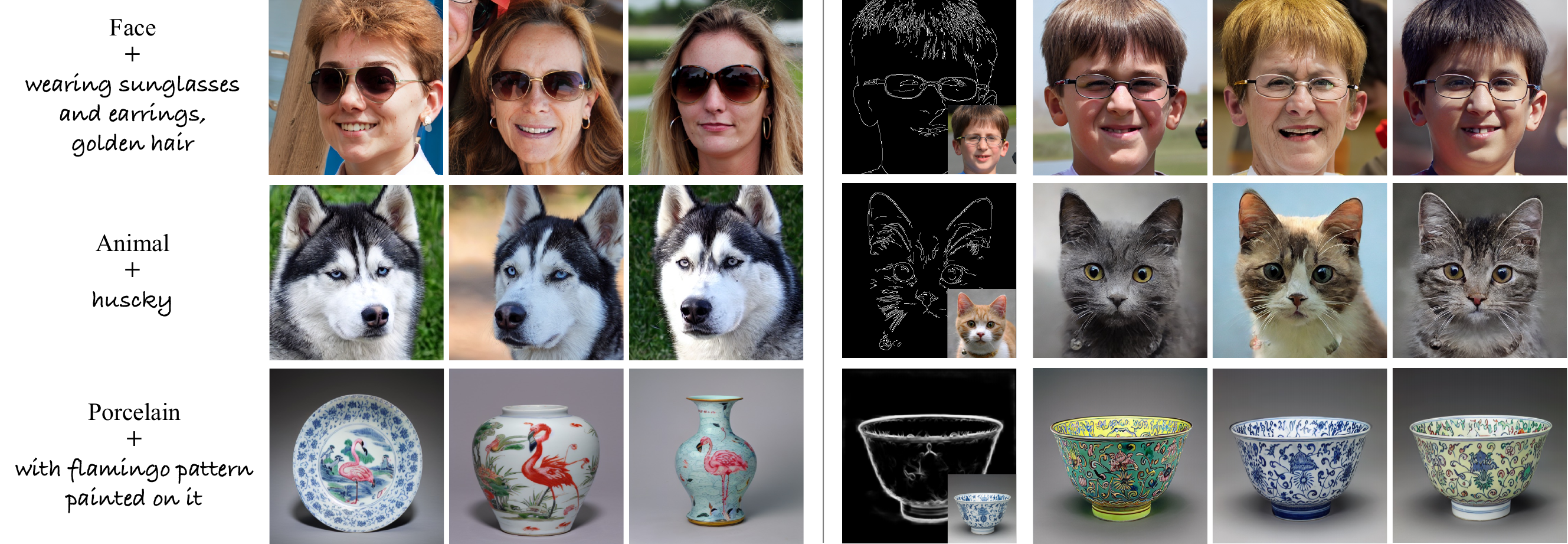}
	\end{center}
	\caption{\textbf{Results of conditional in-domain image generation.} We demonstrate text-conditioned and spatial-conditioned generation results to verify that our proposed methods can effectively preserve the control capability from the pre-trained model. For spatial-conditioned generation, we leverage the off-the-shelf ControlNet model without any extra training.}
	\label{fig:cond}
\end{figure*}


\section{Experiments}
\label{sec:experiments}
\subsection{Experimental Settings}
\label{sec:setting}

\begin{figure*}[t!]
	\begin{center}
		\includegraphics[width=0.8\linewidth]{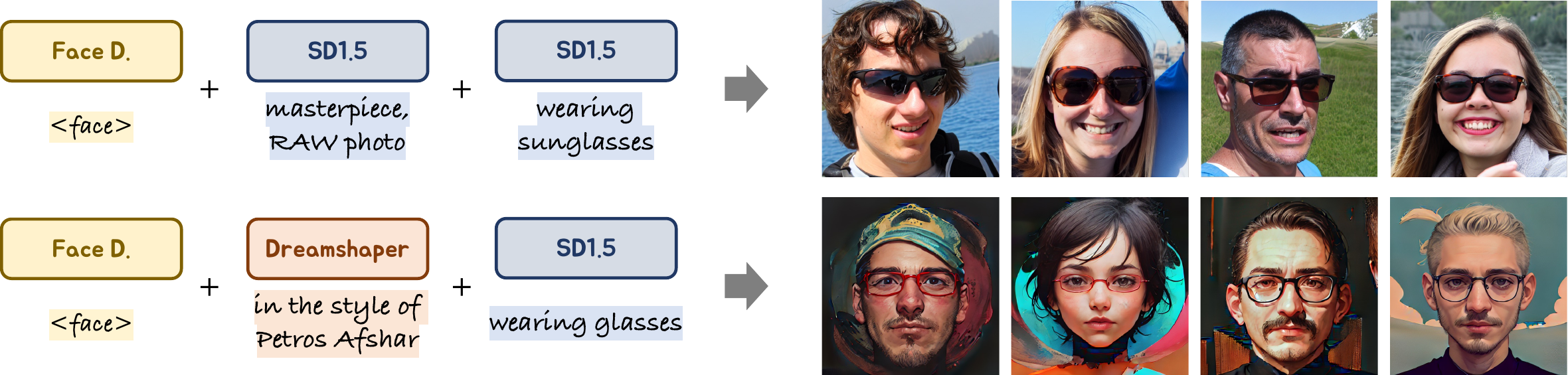}
	\end{center}
	\vspace{-3mm}
	\caption{\textbf{Generative Results under Complex Conditions.} We employ multiple conditions and models to demonstrate the performance of our proposed method in achieving more detailed control over the generation process. }
	\vspace{-2mm}
	\label{fig:multiple_conditions}
\end{figure*}

\begin{figure}[t!]
    \centering
	\includegraphics[width=1\linewidth]{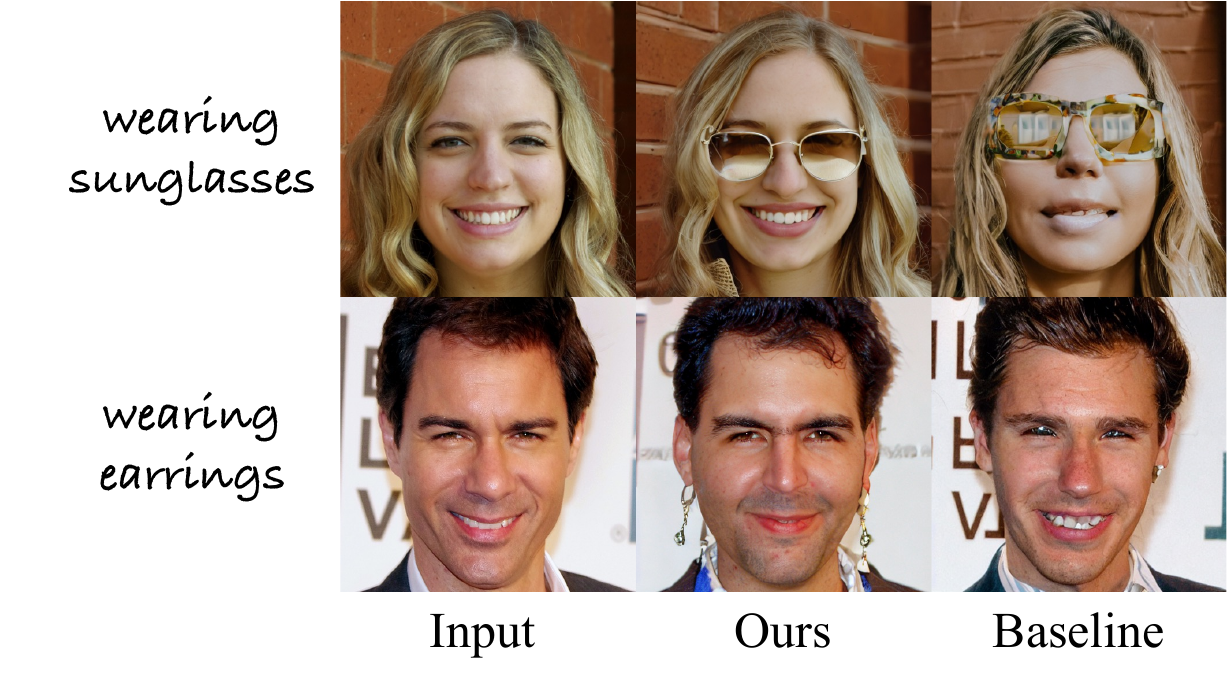}
\vspace{-6mm}
    \caption{\textbf{Image editing on face domain.} We follow SDEdit\cite{meng2021sdedit} to edit facial images equipped with our proposed methods.}
	\label{fig:editing}
\end{figure}

\begin{figure}[t!]
    \centering
	\includegraphics[width=1\linewidth]{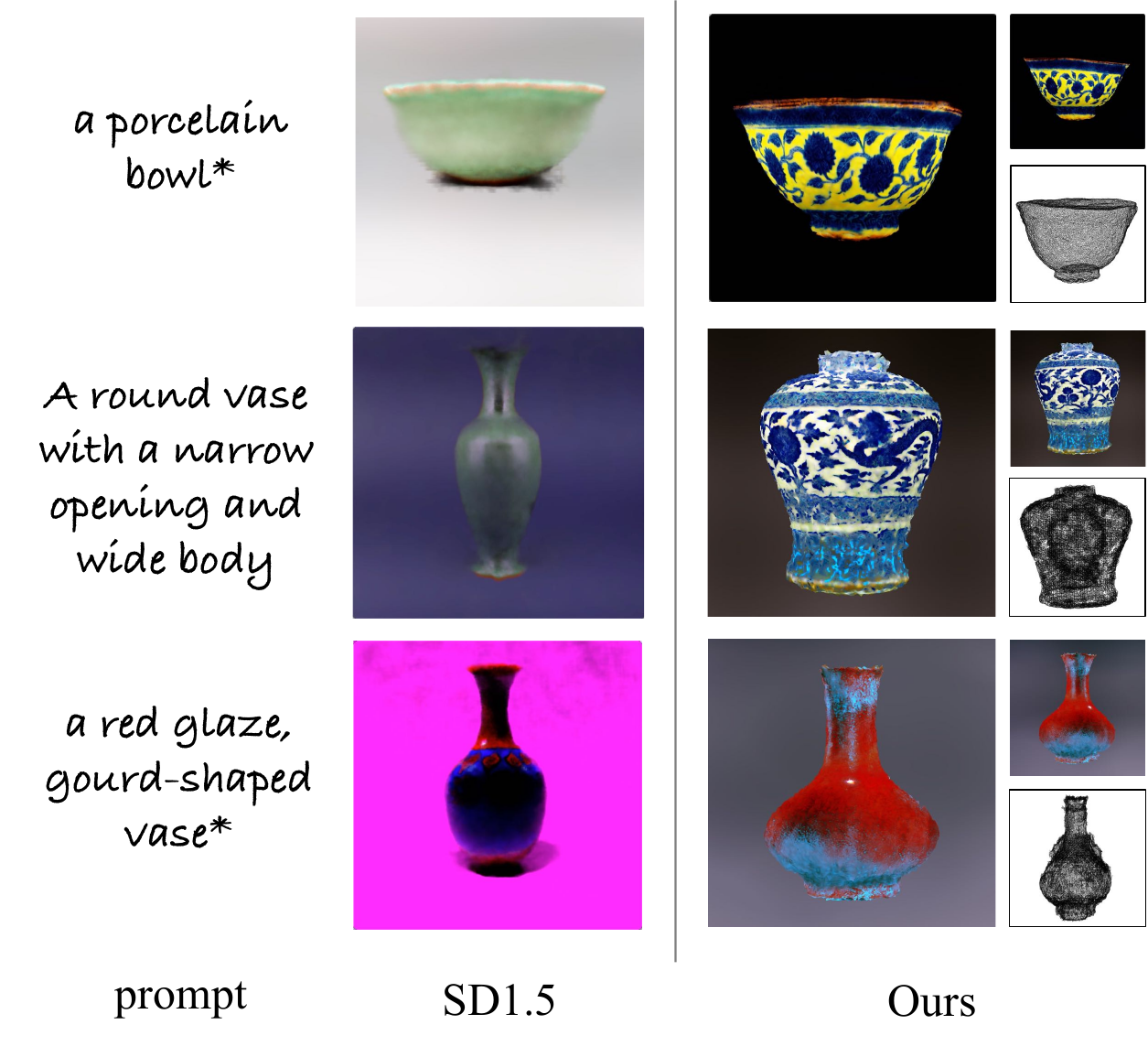}
\vspace{-6mm}
    \caption{\textbf{3D generation on porcelain domain.} We use DreamFusion~\cite{poole2022dreamfusion} based on SD1.5 for comparison. We present two views of the generated objects, including their mesh edges. $*$ denotes the prompt `a DSLR photo of'.}
	\label{fig:3d}
\vspace{-4mm}
\end{figure}

\myparagraph{Datasets}
In this study, we conduct experiments on three domains: face (\emph{i.e.}, FFHQ~\cite{karras2019style}), animal (\emph{i.e.}, AFHQv2~\cite{choi2020starganv2}), and porcelain (\ie a collected dataset), chosen for their diversity and high-resolution (larger than $512\times512$). 
FFHQ is a high-quality human face dataset, containing 70,000 images. Images come from Flickr and are aligned into $1024\times1024$. We further resize images into $512\times512$ without crop.
AFHQv2 is an animal face dataset with a resolution of $512\times512$, containing 15803 images. 
The porcelain dataset contains 5,000 images, collected from the internet.

\myparagraph{Settings} We train the null-text domain diffusion model based on Stable Diffusion v1.5 (SD1.5) at the resolution of 512. For embedding learning phase, 3,200 images are used to attain initial embeddings with batch size of 32 and 100 steps for all domains. For all experiments, we maintain a constant learning rate of 1e-5  with the optimizer of AdamW and a batch size of 32. These experiments are conducted on 8 NVIDIA A100 GPUs, with each running until convergence. All controlling models, including ControlNet and open-sourced customized diffusion models, are borrowed from their official releases based on the corresponding version of Stable Diffusion without any training.
We employ the DPM-Solver++ scheduler~\cite{lu2022dpm} with 100 steps across all models.
For more experimental settings, please refer to Appendix~\ref{sec:apendix_setting}.

\begin{figure*}[t!]
	\begin{center}
		\includegraphics[width=1\linewidth]{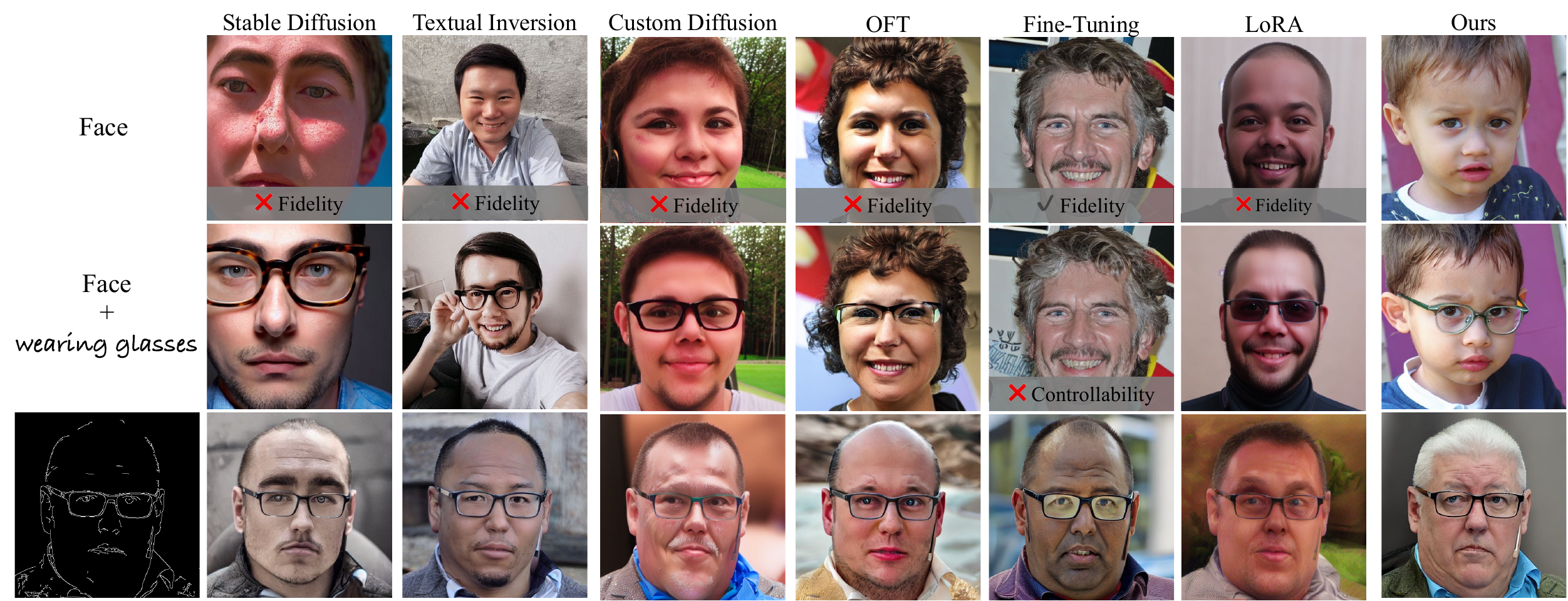}
	\end{center}
	\vspace{-4mm}
	\caption{\textbf{Comparisons to existing training methods.} We evaluate in-domain image generation in unconditional, text-guided, and spatial-guided cases. We fix the random seed to illustrate the controllability for the former two.}
	\vspace{-4mm}
	\label{fig:comparison}
\end{figure*}

\subsection{Results of In-domain Image Generation}
We present the unconditional generation results in \cref{fig:uncond} to illustrate the degree of domain alignment. As shown, our results closely match the distribution of the domain datasets, highlighting the effectiveness of our approach in aligning the model outputs with domain data.
For conditional generation, we showcase the qualitative results in \cref{fig:cond}. We conduct experiments on both text-conditioned and spatial-conditioned generation, utilizing ControlNet models for the latter task. The results clearly demonstrate that our method not only achieves high fidelity with respect to the target domains but also retains the generative controllability seen in pre-trained models, ensuring that the model can be guided by text prompts or spatial conditions as intended. Quantitative results and comparisons can be found in \cref{tab:uncond} and \cref{sec:compare}.

\subsection{Results of Other In-domain Generation Tasks}
Apart from the basic generative tasks, our proposed method also benefits other in-domain generation tasks. In this part, we further demonstrate the effectiveness of our method in a wider range of tasks.

\myparagraph{Image Generation under Complex Conditions}
As discussed in \cref{sec:gcfg}, multiple guidances can be leveraged to control the generation process. In this section, we present the generation results under complex conditions in \cref{fig:multiple_conditions}. Despite the introduction of complex conditions, the generated images maintain high fidelity to the target domain, demonstrating both the robustness and quality of our method. Additionally, our generative framework allows us to leverage additional open-source models for controlling the generation process, such as using DreamShaper to control the style.

\myparagraph{Image Editing} In \cref{fig:editing}, we implement SDEdit\cite{meng2021sdedit} on the face domain, where the input image is first encoded into latent space and added noise and then we use a similar generation pipeline presented in \cref{sec:pipeline}. Detailed settings can be found in \cref{sec:setting_edit}. Our results are more aligned with the distribution of real human face images, demonstrating better visual quality.

\myparagraph{Text-to-3D Generation} In \cref{fig:3d}, we leverage our model on porcelain domain for text-to-3d porcelain generation and compare it with classic text-to-3d model~\cite{poole2022dreamfusion} on SD1.5. Specifically, we begin by generating porcelain images based on the provided prompt, which are then converted into 3D models using an off-the-shelf image-to-3D method~\cite{liu2023zero}. Subsequently, we fine-tune the resulting 3D models using our porcelain model, guided by the SDS~\cite{poole2022dreamfusion} loss. Lastly, we directly convert the NeRF models into meshes. Detailed settings and additional ablation studies on 3D generation can be found in \cref{sec:setting_3d}. Our results demonstrate superiority in color richness, fine detail, and alignment with the dataset.

\begin{figure*}[t!]
\centering
	\subfloat[In-domain Generation]{
        \includegraphics[width=0.3\textwidth]{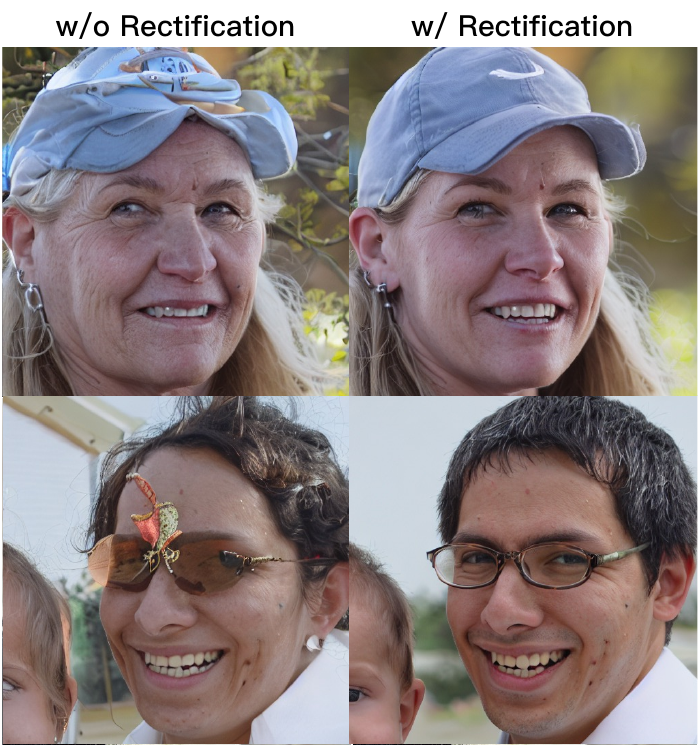}
        \label{fig:ablation_uncond}
    }\hspace{1mm}
	\subfloat[Open-Sourced Diffusion Models]{
        \includegraphics[width=0.64\textwidth]{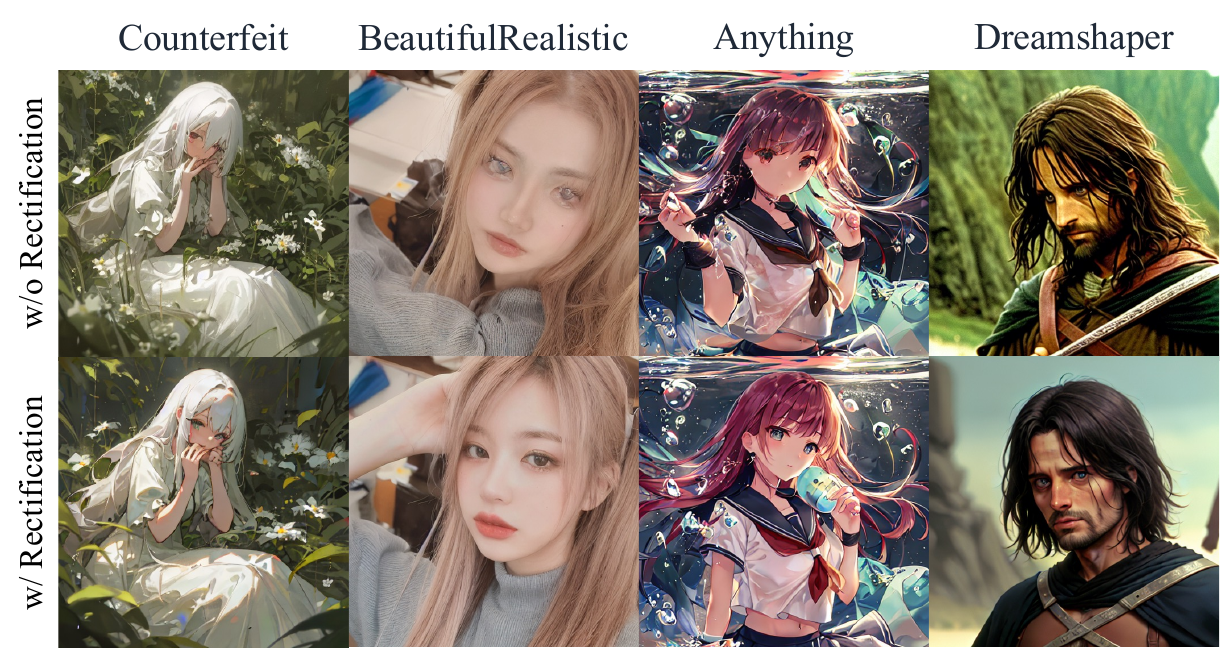}
        \label{fig:civitai_gcfg}
    }
	\caption{\textbf{Effects of Unconditional Guidance Rectification.} (a) Effects in Facial Domain Generation. 
(b) Unconditional guidance rectification also improves the open-sourced diffusion models.
}
\vspace{-2mm}
	\label{fig:ablation}
\end{figure*}

\begin{table*}[t]
\centering
\caption{\textbf{Quantitative results of unconditional generation} We evaluate FID and human preference on two datasets. Our method achieves excellent generation results among baselines.$^{\dagger}$: reproduced results in $512\times 512$ resolution. \textbf{Bold}: the best result. \underline{Underline}: the second best result.}
\resizebox{1.0\textwidth}{!}{%
\begin{tabular}{l|c|cccccc|cccc}
\toprule
\multirow{3}{*}{Method} & \multirow{3}{*}{\#param.} & \multicolumn{6}{c|}{Unconditional Generation} & \multicolumn{4}{c}{Conditional Generation} \\ \cline{3-12}
& & \multicolumn{2}{c}{Face} & \multicolumn{2}{c}{Animal} & \multicolumn{2}{c|}{Porcelain} & \multicolumn{2}{c}{Text-Guided} & \multicolumn{2}{c}{Spatial-Guided} \\
 & & FID$\downarrow$ & Pref.$\uparrow$ & FID$\downarrow$ & Pref.$\uparrow$  & FID$\downarrow$ & Pref.$\uparrow$ & Align.$\uparrow$ & Pref.$\uparrow$ & Align.$\downarrow$ & Pref.$\uparrow$\\\hline
Textual Inversion~\cite{gal2022image} &  $<1$K &   70.62       &   0.3\% &   44.75    &   2.5\% & 112.76 & 1.7\% & \textbf{0.97} & 0.2\% & 0.22 & 1.5\%\\
Custom Diffusion~\cite{kumari2023multi} & 44M &   40.98      &    1.6\%   &   49.98 &     1.8\% & 98.77 & 0.6\% &   0.74 &     1.4\% &    \textbf{0.20}    &      1.7\%      \\
OFT~\cite{qiu2023controlling} & 23M &   27.50      &    2.9\%   &   38.74 &     1.4\%   & 68.84 & 1.2\% &   0.77 &    3.2\% &    \underline{0.21}    &      \underline{9.6\%}     \\
Fine-tuning & 859M &  \underline{12.37}      &    \underline{6.4\%} &    \underline{23.92}  &     \underline{12.4\%} & \underline{62.76} & \underline{10.8\%} &   0.24 &    4.5\% &    0.22 &    4.9\%\\
Fine-tuning + LoRA\cite{hu2021lora} & 0.19M &  23.76      &    {1.3\%} &    24.19  &     1.7\% & 73.46 & 0.3\% &   0.76 &    \underline{6.7\%} &    0.22 &    9.1\%\\
Ours   & 810M & \textbf{6.57}    &   \textbf{87.5\%}      &   \textbf{18.82}  &     \textbf{80.2\%}   & \textbf{56.46} & \textbf{85.4\%} &   \underline{0.89} & \textbf{84.0\%} &    \textbf{0.20} &    \textbf{73.2\%}   \\  
\bottomrule
\end{tabular}%
}
\vspace{-4mm}
\label{tab:uncond}
\end{table*}

\subsection{Comparisons to Fine-tuning Techniques}
\label{sec:compare}
To validate the superiority of our proposed method, we compare our method with several fine-tuning baselines, including vanilla fine-tuning, parameter-efficient fine-tuning, and some training techniques borrowed from other tasks. 
For detailed experimental settings, please refer to Appendix\ref{sec:baseline_setting}.

\myparagraph{Baselines} 
To establish comprehensive baselines, we conduct experiments on several training techniques, including vanilla fine-tuning, low-rank fine-tuning~\cite{hu2021lora}, textual inversion~\cite{gal2022image}, custom diffusion~\cite{kumari2023multi}, and OFT mechanism~\cite{qiu2023controlling}. 
To train diffusion models without text annotation, we adopt prompt templates, utilizing formats such as \texttt{a photo of [V] <domain name>}. These additional tokens \texttt{[V]} are omitted in compared methods where text embedding is not trained.

\myparagraph{Evaluation Setting} For unconditional generation, we calculate FID on 50,000 images. For text-guided generation, we generate 1,000 images over 10 attribute-related prompts on the facial domain and employ facial predictors to ascertain whether these attributes are accurately represented in the generated images. Besides, we sample 200 images from CelebA-HQ dataset to extract canny information as the spatial condition and compute the canny discrepancy. We also report human preference (denoted as `Pref') by collecting the win rates of generative images across all comparisons. Detailed settings are presented in \cref{sec:baseline_setting}.

\myparagraph{Qualitative Comparisons} 
We conducted a qualitative comparison of our method against previous training techniques, as depicted in Fig~\ref{fig:comparison}. Methods like textual inversion, custom diffusion, and LoRA, which adhere to a parameter-efficient form, exhibit noticeably lower fidelity compared to others. Vanilla fine-tuning, in particular, demonstrates a decline in controllability during extended training periods. 
This results in a diminished capacity for text-guided generation and the emergence of unnatural artifacts, especially when used in conjunction with ControlNet.
In contrast, our method excels in both fidelity and control capabilities.

\myparagraph{Quantitative Comparisons}
We conduct the evaluation of unconditional generation and report the results in Tab~\ref{tab:uncond}. Our approach demonstrates state-of-the-art generation quality, evidenced by a significantly lower FID compared to baseline methods and the highest human preference rating. 
For text-guided generation, Textual Inversion shows a higher alignment with the text prompts than our method. However, this is achieved at the cost of image quality, owing to its reliance on a very limited set of learnable parameters. This compromise in quality is clearly evident in the comparisons presented in Fig~\ref{fig:comparison}. In contrast, in spatial-guided generation, where semantic features are integrated into the UNet, the alignment degree among different methods is similar. Notably, our results not only align well with the control signals but also achieve the highest ratings in human preference assessments.


\subsection{Effects of Unconditional Guidance Rectification} 
\label{sec:discussion}
In previous methods, unconditional guidance is predicted by the customized model, leading to inaccurate estimation of probability in sampling. In our framework, the unconditional guidance is rectified by straightforwardly predicting by diffusion priors. We compare generation results in Fig~\ref{fig:ablation_uncond}. Our method significantly improves image quality, generating fewer artifacts and attaining photorealistic results. Besides, we utilize four widely-used open-sourced models. 
The comparative results are presented in Fig~\ref{fig:civitai_gcfg}. 
Without rectification, the generation quality shows slight degradation, as evidenced by less detailed eyes and less realistic hair textures. The implementation of guidance decoupling substantially enhances generative performance via the rectification of unconditional guidance.


\section{Conclusion}

In this research, we address the task of enhancing large-scale diffusion models for in-domain generation using only image data. Our goal is to enable these models to generate images that faithfully represent the given domains, achieving high fidelity, diversity, and controllability. Moreover, our approach is designed to be compatible with various control methods, allowing the model to perform a wide range of generative tasks.
To tackle this challenge, we propose a guidance-decoupled prior preservation mechanism that separates conditional guidance into two components: domain guidance for alignment and control guidance for maintaining the models controllability. Additionally, we introduce an efficient learning approach by incorporating a null-text diffusion model, ensuring both simplicity and effectiveness. We then outline the generative pipeline of our method to achieve diverse in-domain generation tasks.
We conduct experiments across three domains to demonstrate the effectiveness of our proposed method.

\newpage
\section{Acknowledgements}
This work was supported by the Young Scientists Fund of NSFC (Grant No. 62406035).

{
    \small
    \bibliographystyle{ieeenat_fullname}
    \bibliography{main}
}

\clearpage

\appendix

\section{Analysis of Condition Independence}
\label{sec:appendix_independent}

Although integration of multiple guidances in the denoising process is popular recently in many tasks\cite{}, independence of conditions is ignored now. We first provide the total proof of \cref{eq:gcfg}.
\begin{property}
Denoting a sets of independent conditions as $C=\{c_1,\cdots,c_k\}$ with manually defined intensities $w=\{w_1,\cdots,w_k \}$, we can use multiple UNets $\epsilon_\theta$ to predict conditional and unconditional: $\epsilon_{\theta_0}(z_t)=\nabla_z\log p(z_{t})$ and $\epsilon_{\theta_i}(z_t,c_i)=\nabla_z\log p(z_{t}|c_i)$. The reverse process in each timestamp is as follows:
\begin{align}
\label{eq:gcfg}
\hat\epsilon(z_t,C)=\epsilon_{\theta_0}(z_t)+\sum_{i=1}^K w_i(\epsilon_{\theta_i}(z_t,c_i)-\epsilon_{\theta_0}(z_t))
\end{align}
\end{property}

\begin{proof}
Since $C=\{c_1,\cdots,c_k\}$ are independent, we have $p(C|z_{t+1})=\prod p(c_i|z_{t+1})^{w_i}$.
The denoising process is then converted into:
\begin{align}
&\nabla_{z_t}\log p(z_{t})+\nabla_{z_t}\log p(C|z_{t}) \nonumber \\
=&\nabla_{z_t}\log p(z_{t})+\sum_{i=1}^Kw_i\nabla_{z_t}\log p(c_i|z_{t})\nonumber\\ 
=&\nabla_{z_t}\log p(z_{t})+\sum_{i=1}^Kw_i\nabla_{z_t}\log \frac{p(z_{t}|c_i)}{p(z_{t})} \nonumber\\
=&\nabla_{z_t}\log p(z_{t})+\sum_{i=1}^Kw_i(\nabla_{z_t}\log p(z_{t}|c_i)-\nabla_{z_t}\log{p(z_{t})}) \nonumber
\end{align}
In Stable Diffusion, we use a UNet $\epsilon$ to predict each term: $\epsilon(z_t)=\nabla_z\log p(z_{t})$ and $\epsilon(z_t,c_i)=\nabla_z\log p(z_{t}|c_i)$.
\end{proof}

\begin{figure}[h!]
	\begin{center}
		\includegraphics[width=0.7\linewidth]{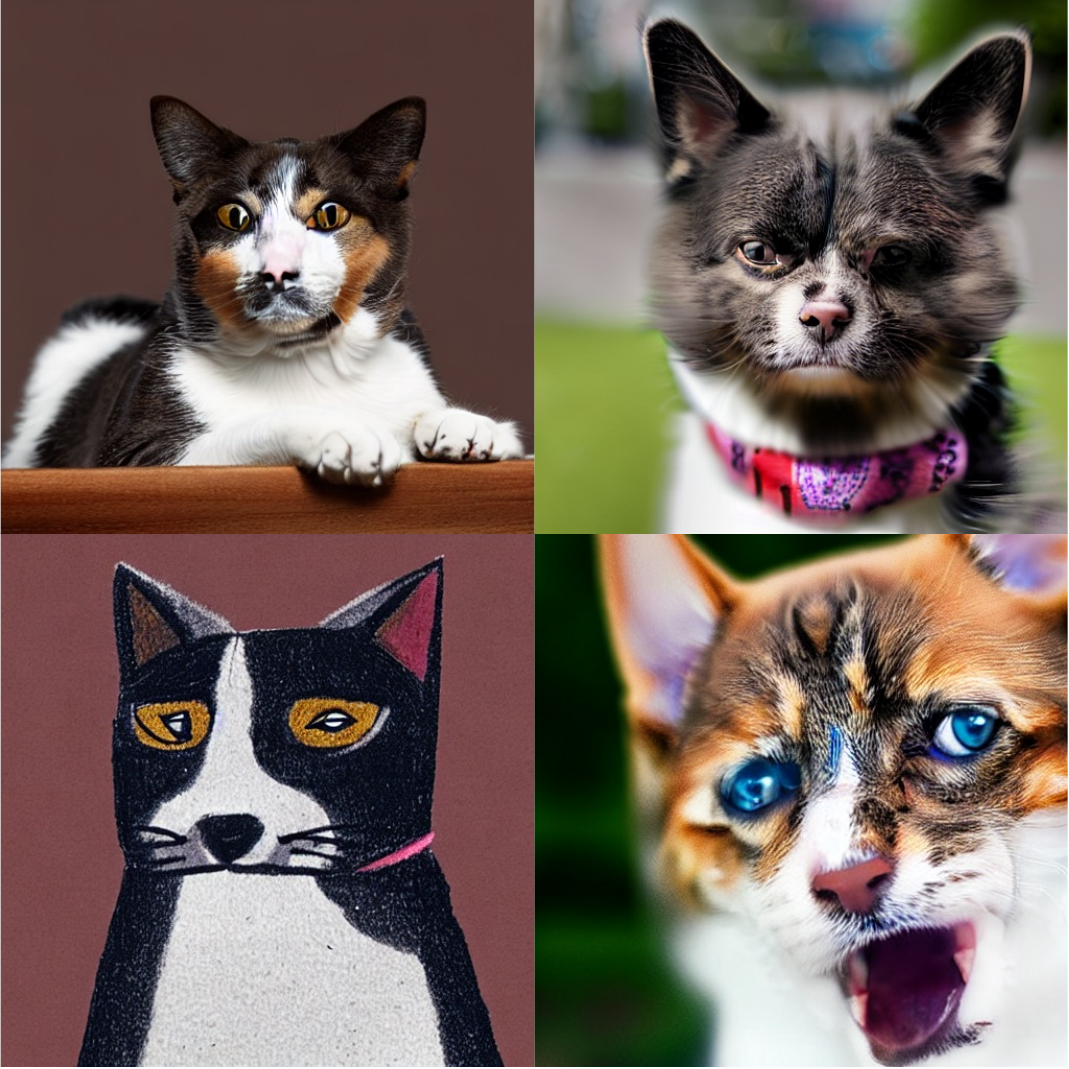}
	\end{center}
	\caption{\textbf{Example to reveal condition independence.} We use two text conditions \texttt{a photo of a cat} and \texttt{a photo of a dog}. The generative results represent a mixture creature of dog and cat.}
	\label{fig:dog}
\end{figure}

An important assumption is the independence of each condition.
Here we provide a simple example to reveal the condition independence and illustrate how we can utilize multiple conditions in in-domain generation. We use two text conditions $c_1=$\texttt{a photo of a cat} and $c_2=$\texttt{a photo of a dog}, and attain the guidance by: $\hat\epsilon=(1-2w)\epsilon(z_t)+w\cdot\epsilon(z_t,c_1)+w\cdot\epsilon(z_t,c_2)$. The expected result, which was to produce one cat and one dog, did not occur; instead, a hybrid creature, combining features of both a cat and a dog, was generated as can be seen in \cref{fig:dog}.

The theoretical explanation for this phenomenon is that during the denoising process, the denoising direction is independently guided by each component to maximize the likelihood of each condition independently. It means that the generative result achieves the high probability of both $p(c_1|x)$ and $p(c_2|x)$. Hence, an effective way to utilize multiple guidance for synthesis is by using conditions which can represent the whole image. For example, in text-guided face-domain generation, using \texttt{a photo of face, wearing glasses} would be much better than using \texttt{wearing glasses}.

\section{Condition Decoupling}
\begin{figure}[h!]
	\begin{center}
		\includegraphics[width=1.0\linewidth]{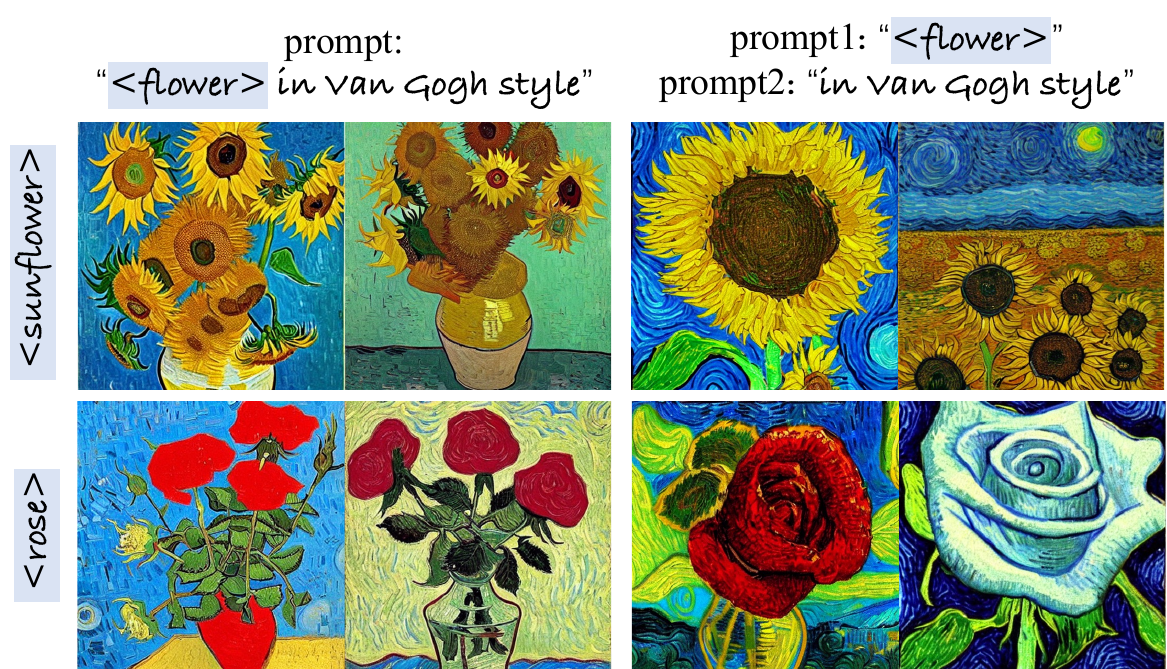}
	\end{center}
	\caption{\textbf{Example of condition decoupling.} Decouple conditions can make contents independent. It helps mitigate the bias in diffusion priors, leading to more diverse results.}
	\label{fig:vangogh}
\end{figure}

Apart from domain guidance and control guidance, we further demonstrate a general condition decoupling technique. It can also explain that we could using our in-domain diffusion model to generate out-of-domain results (like stylization generation). 
Using multiple guidances also has the unique ability to decouple relationships between contents that are typically related in the real world, thereby enabling the generation of more diverse images. For instance, if the goal is to generate images of a sunflower in the style of Van Gogh, as opposed to replicating Van Gogh's specific sunflower paintings, we can apply two conditions: \texttt{sunflower} and \texttt{in Van Gogh style}. This distinction is showcased in \cref{fig:vangogh}. Using the combined prompt \texttt{sunflower in Van Gogh Style} tends to produce images closely resembling Van Gogh's paintings. In contrast, employing decoupled conditions results in a much broader diversity of images. This method proves equally effective for generating concepts similar to those in the training dataset, such as \texttt{rose}.

\begin{figure}[t!]
    \centering
	\includegraphics[width=1\linewidth]{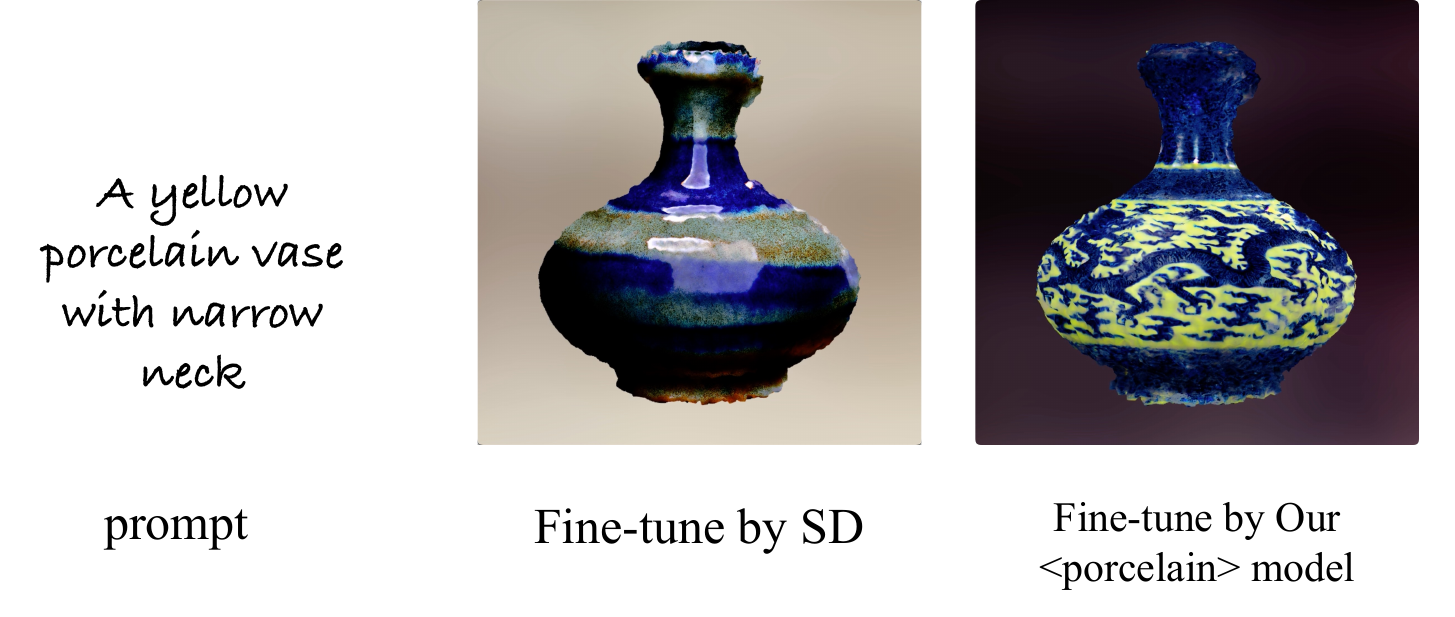}
    \caption{\textbf{Ablation study on 3D generation.} During the 3D fine-tuning process, we replaced our porcelain model with basic SD, resulting in a noticeable decline in generation quality. The distinctive porcelain patterns in the output disappeared completely. }
	\label{fig:3d_ablation_appendix}
\end{figure}

\section{Experimental Details}
\label{sec:apendix_setting}
\subsection{Inference and Evaluation} 
We employ the DPM-Solver++ scheduler~\cite{lu2022dpm} with 100 steps across all models while using 25 steps for stylized results to achieve better results. Regarding the guidance scale, we perform grid searches on all models and conditions, with an interval of $0.25$. In text-guided generation, we combine the prompts for baselines, like \texttt{a photo of <V> face, wearing glasses}.
To quantitatively measure the fidelity and controllability of the generated images, we incorporate the FID in unconditional generation, and alignment in conditional generation. Moreover, we also undertake a user study to gather human preferences, providing a qualitative dimension to our evaluation criteria and ensuring a holistic view of the performance of the methods under scrutiny

\myparagraph{Text-guided Generation} Although the measure of text-image similarity, often quantified by CLIP scores, is a common metric for evaluating the alignment in text-guided image generation, we observed its limitations in accurately reflecting facial attributes. For instance, in the case of the prompt `wearing glasses,' the difference in CLIP scores between images with and without glasses is relatively marginal, often around a value of 1 (e.g., 18 to 19). Due to this lack of pronounced differentiation, we adopt an alternative approach for evaluation.

\begin{table}[h]
\caption{\textbf{Prompts used for text-guided generation evaluation.} We generate 100 images of each prompt for evaluation.}
\centering
\resizebox{0.5\textwidth}{!}
{
\begin{tabular}{c|c}
\toprule
\texttt{wearing glasses} & \texttt{wearing sunglasses} \\ \hline
\texttt{wearing hat} & \texttt{smilling} \\ \hline
\texttt{male} & {\texttt{female}} \\ \hline
\texttt{white people} & \texttt{black people} \\ \hline
\texttt{asian people} & \texttt{square face} \\
\bottomrule
\end{tabular}
}
\label{tab:eval_prompt}
\end{table}

We utilize a set of attribute-specific prompts, detailed in Tab~\ref{tab:eval_prompt}, to generate images. Subsequently, we employ facial attribute predictors to ascertain whether the specified attribute (e.g., glasses) is accurately represented in the generated images. The effectiveness of our method is then quantified by the ratio of successful samples where the controlled attribute is correctly manifested in the results.

\subsection{Fine-Tuning Baselines}
\label{sec:baseline_setting}
To establish a comprehensive baseline, we employ several state-of-the-art fine-tuning methods: Textual Inversion~\cite{gal2022image}, DreamBooth~\cite{ruiz2023DreamBooth}, Custom Diffusion~\cite{kumari2023multi}, and OFT~\cite{qiu2023controlling}, along with our method. 
Textual Inversion and Custom Diffusion serve as parameter-efficient comparisons, streamlining the experimental setup, while OFT is included for its regularization properties. DreamBooth fine-tunes a concept token \texttt{<V>} and UNet's parameters without class-specific prior preservation loss since it is not compatible to customize concept.
The experimental distinction between baselines and our concept-centric diffusion models lies in the use of text prompts. We adopt prompt templates inspired by DreamBooth \cite{ruiz2023DreamBooth}, utilizing formats such as \texttt{a photo of <V> face}. These additional tokens \texttt{<V>} are omitted in compared methods where text embedding is not trained

\myparagraph{Spatial-guided Generation} To assess the effectiveness of spatial-guided generation, we select a sample of 200 images from the CelebA-HQ dataset. For each image, we generate corresponding canny images to use as conditions. We then produce 5 generated results per condition and compute the discrepancy between the canny conditions and the corresponding generative results, providing a quantitative measure of the alignment accuracy.

\myparagraph{User Study} To gauge human preferences in both unconditional and conditional generation contexts, we organize a user study. We gain generative results from all methods using the same random seed across all methods and present these images to participants in a random sequence. Participants are instructed to choose the best image from the given set. We record the frequency with which each method is selected as the best, and this data is used to calculate the win rate for each method. The win rates serve as an indicator of human preference (denoted as 'Pref.') and are presented in the main paper.


\subsection{Other In-domain Generation Settings} 
\subsubsection{Image Editing}
\label{sec:setting_edit} 
We following SDEdit to edit facial images with diffusers implementation\footnote{\url{https://github.com/huggingface/diffusers/blob/main/src/diffusers/pipelines/stable_diffusion/pipeline_stable_diffusion_img2img.py}}. We use noising strength of 0.6 and inference steps of 20. We follow \cref{sec:pipeline} to adjust the guidance scales. The editing process is the same as text-guided in-domain generation, we utilize original SD1.5 to predict unconditional guidance and text-conditioned guidance and use the trained facial domain diffusion model to predict domain guidance.

\subsubsection{Text-to-3D Generation}
\label{sec:setting_3d}
Our design involves three stages to generate a 3D porcelain. First, we use our model to create a 2D porcelain image. Next, we employ the classic Zero-1-to-3 model for image-to-3D conversion to obtain a rough target 3D model. Finally, we fine-tune this model using our porcelain model and SDS loss. During the fine-tuning process, we set the CFG parameters to 100 on SD and 50 on our porcelain model. Additional results is shown on Fig~\ref{fig:3d_appendix}. Our implementation is based on stable-dreamfusion\footnote{\url{https://github.com/ashawkey/stable-dreamfusion}}.

\noindent\textbf{Effects of our porcelain model in 3D generation.} During the 3D
fine-tuning process, we replaced our porcelain model with basic SD, resulting in a noticeable decline in generation quality. The distinctive porcelain patterns in the output disappeared completely, as shown in Fig~\ref{fig:3d_ablation_appendix}.

\section{More Qualitative Results}

\subsection{Unconditional Concept-centric Generation}
\begin{figure*}[t!]
	\begin{center}
		\includegraphics[width=1.0\linewidth]{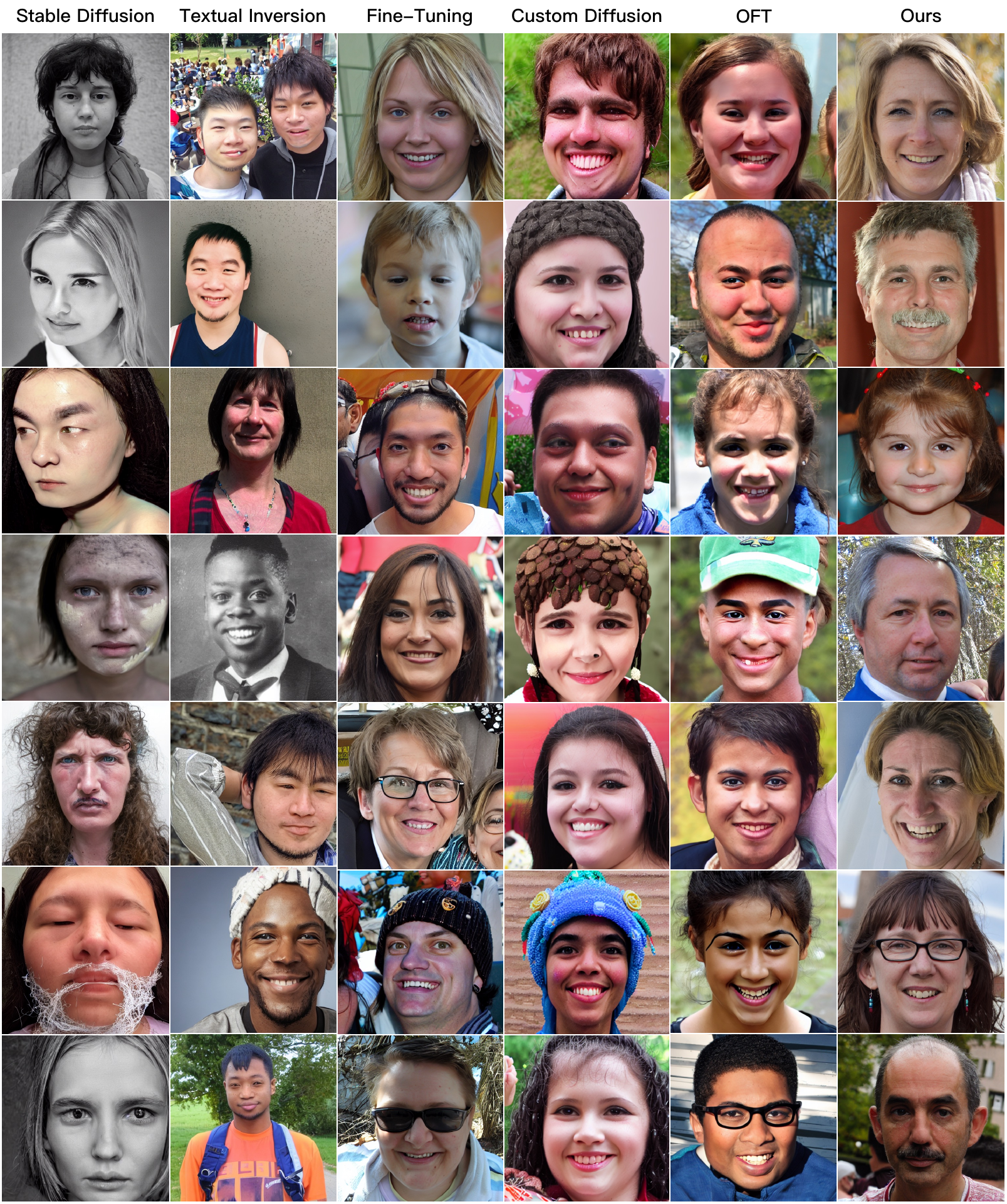}
	\end{center}
	\vspace{-2mm}
	\caption{\textbf{Unconditional generation results on FFHQ.} We illustrate the unconditional results of all models trained from our method and baselines.}
	\vspace{-2mm}
	\label{fig:appendix_uncond_ffhq}
\end{figure*}

\begin{figure*}[t!]
	\begin{center}
		\includegraphics[width=1.0\linewidth]{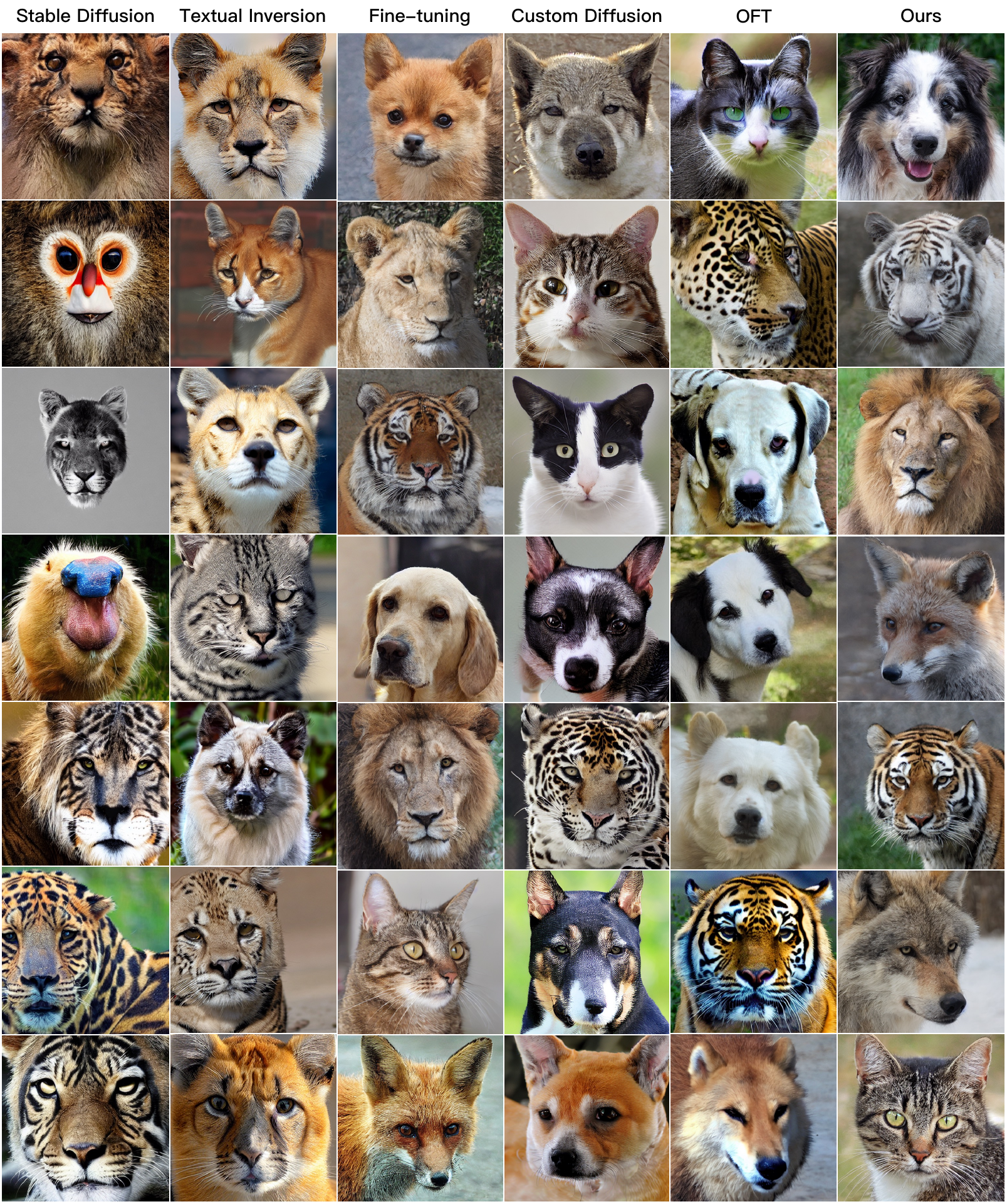}
	\end{center}
	\vspace{-2mm}
	\caption{\textbf{Unconditional generation results on AFHQv2.} We illustrate the unconditional results of all models trained from our method and baselines.}
	\vspace{-2mm}
	\label{fig:appendix_uncond_afhq}
\end{figure*}

\begin{itemize}
\item Fig~\ref{fig:appendix_uncond_ffhq} illustrates the comparison of unconditional generative results on FFHQ.
\item Fig~\ref{fig:appendix_uncond_afhq} illustrates the comparison of unconditional generative results on AFHQv2.
\end{itemize}

\subsection{Conditional Generation}
\begin{figure*}[t!]
	\begin{center}
		\includegraphics[width=1.\linewidth]{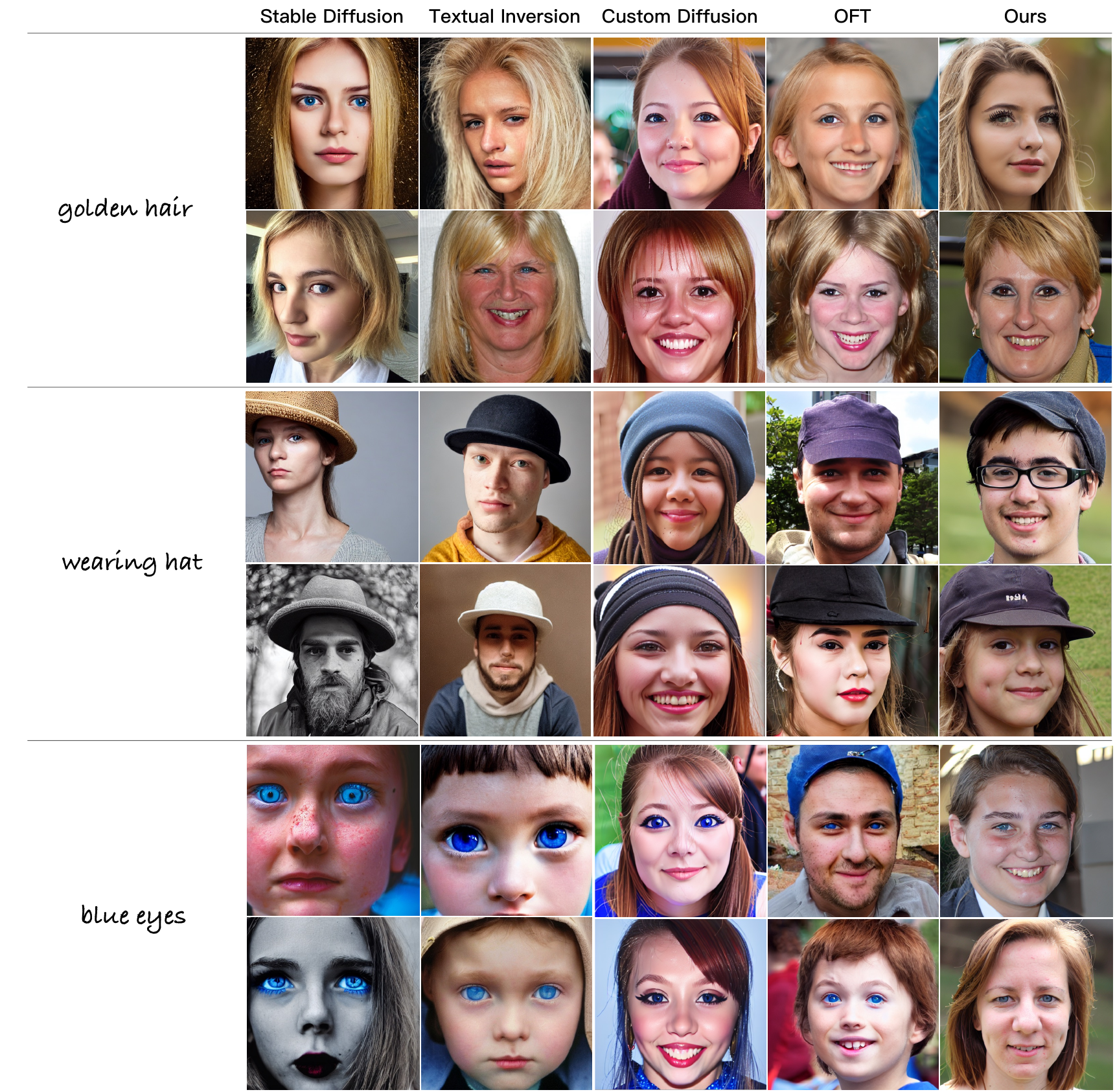}
	\end{center}
	\vspace{-2mm}
	\caption{\textbf{Text-guided generation comparison across baselines and our method.} Since fine-tuning with large-scale training process almost loses controllability (as shown in Fig~\ref{fig:comparison}), we evaluate other methods in this part.}
	\vspace{-2mm}
	\label{fig:appendix_comparison_cond}
\end{figure*}

Fig~\ref{fig:appendix_comparison_cond} illustrates the comparison of text-guided generative results on FFHQ.

\begin{figure*}[t!]
	\begin{center}
		\includegraphics[width=1\linewidth]{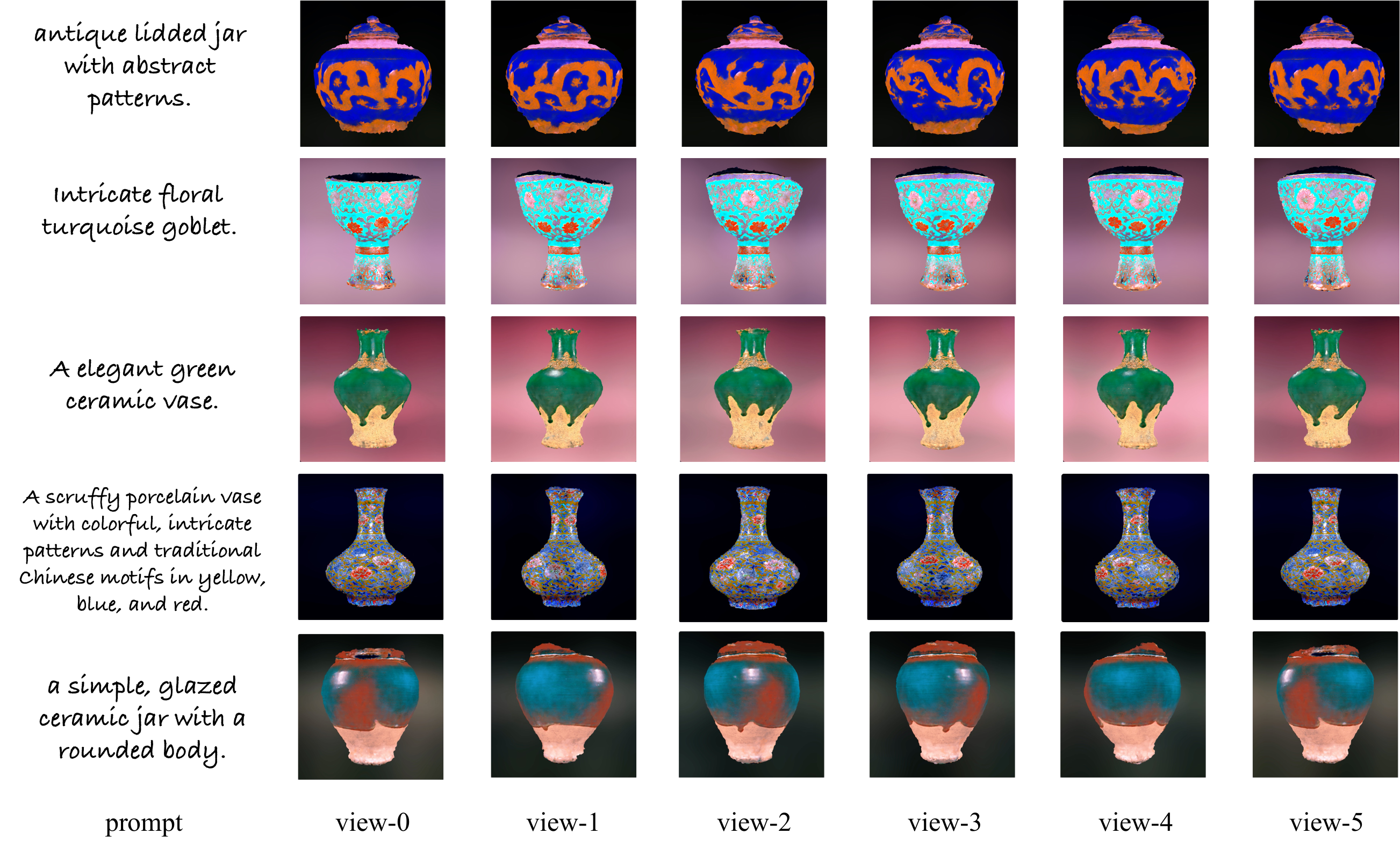}
	\end{center}
	\caption{\textbf{Additional results of 3D generation with porcelain model.}}
	\label{fig:3d_appendix}
\end{figure*}

Fig~\ref{fig:3d_appendix} illustrates the results of 3D generation within porcelain domain.

\section{Limitations}
\label{sec:limitations}
Using multiple models to estimate different guidances slightly increases both inference time and memory usage. Originally, the generation process required predicting two guidances, but in-domain generation requires predicting three guidances, resulting in a 50\% increase in computation time. To accelerate this process, we utilized multiple GPUs by placing the concept diffusion model and the original diffusion model on separate GPUs for parallel computation, which only increased the generation time by approximately 16\%. Regarding memory usage, incorporating an additional concept diffusion model increases the memory requirement for 512-resolution generation from 3.5GB to 5GB, which is acceptable for most GPU devices. One foreseeable solution to the computation time and memory issues is to train on a distilled, smaller version of Stable Diffusion, as learning domain guidance does not require a large UNet. We plan to explore this in future work.
Meanwhile,  text prompts can not be too much out-of-domain, which causes conflict between domain guidance and control guidance.

\end{document}